%% file: paper.tex
\pdfoutput=1

\documentclass[letterpaper, 10pt, journal]{IEEEtran}      

\IEEEoverridecommandlockouts                              

\usepackage{multicol}
\usepackage{multirow}
\usepackage{graphicx}
\usepackage{amsmath}
\usepackage{algpseudocode}
\usepackage{color}
\usepackage{xcolor}
\usepackage{graphicx}
\usepackage{bm}
\usepackage{amsmath}
\usepackage{amssymb}
\usepackage{tabularx}
\usepackage{subcaption}
\usepackage{url}
\usepackage{cite}
\usepackage{prettyref}
\usepackage{booktabs}
\usepackage{siunitx}
\usepackage{comment}
\usepackage{booktabs}
\usepackage{hyperref}
\usepackage{tabularx, booktabs}
\usepackage{makecell} 
\usepackage[table]{xcolor} 
\usepackage{pifont} 
\usepackage{tabularx} 
\definecolor{Gray}{gray}{0.9}

\usepackage[all]{hypcap}
\usepackage{tabularx} 

\usepackage{adjustbox} 

\usepackage[ruled]{algorithm2e} 

\SetAlFnt{\small}
\SetAlCapFnt{\small}
\SetAlCapNameFnt{\small}
\SetAlCapHSkip{0pt}

\newrefformat{fig}{Figure~\ref{#1}}
\newrefformat{sec}{Section~\ref{#1}}
\newrefformat{tab}{Table~\ref{#1}}
\newrefformat{alg}{Algorithm~\ref{#1}}
\newrefformat{eqn}{Equation~\ref{#1}}

\hypersetup{
    colorlinks,
    linkcolor={blue!80!black},
    citecolor={blue!80!black},
    urlcolor={blue!100!black}
}

\makeatletter
\newcommand{\printfnsymbol}[1]{%
  \textsuperscript{\@fnsymbol{#1}}%
}
\makeatother

\title{ \LARGE \bf Reduced-order Neural Modeling with Differentiable Simulation for \\ High-Detail  Tactile Perception 
}

\author{
    Yuhu Guo$^{1}$,
    Zhikai Shen$^{2}$,
    Jiasheng Qu$^{2}$,
    Chenghao Qian$^{3}$,
    Yuming Huang$^{1}$,
    Bin Chen$^{4*}$,
    Guoxing Fang$^{2*}$%
    \thanks{
        $^{1}$The University of Manchester, UK 
        $^{2}$The Chinese University of Hong Kong, Hong Kong SAR, China 
        $^{3}$University of Leeds, UK 
        $^{4}$University of Melbourne, Australia 
        $^*$Corresponding Author( bin.chen@unimelb.edu.au; guoxinfang@mae.cuhk.edu.hk)
        Project Page: \href{https://anabur920.github.io/tac2detail.github.io/}{tac2detail.github.io}
    }
}

\begin{document}
\maketitle

\begin{abstract}
Tactile perception is key to dexterous manipulation, yet simulating high-resolution elastomer deformation remains computationally prohibitive. Finite element methods (FEM) deliver high fidelity but demand costly remeshing, while Material Point Methods (MPM) suffer from heavy particle-memory tradeoffs.  We propose a {reduced-order neural simulation framework} that couples coarse-grained MPM dynamics with an implicit neural decoder to reconstruct sub-particle tactile details from compact latent states. 
The framework learns a continuous deformation manifold from paired high- and low-resolution simulations, enabling physically consistent, differentiable inference. 
Compared to the TacIPC, our method achieves over 65\% faster simulation and {40\% lower memory usage}, while maintaining better geometric fidelity. 
In tactile rendering and 3D surface reconstruction, our methods further improve accuracy by 25\% and produce realistic depth images and surface mesh within a faster inference speed. 
These results demonstrate that the proposed reduced-order neural model enables high-detail, physically grounded tactile simulation with substantial efficiency gains for robotic interaction and optimization. 

\end{abstract}


\input{tex/Intro}

\input{tex/Related_work}

\input{tex/Pipeline_Overview}
\input{tex/Experiments}

\input{tex/secConclusion}



{\small
\bibliographystyle{IEEEtran}
\bibliography{references}
}

\end{document}

%% file: tex/Intro.tex
\section{Introduction}
Tactile perception enables the recovery of rich geometric and physical cues from object contact by capturing surface deformation. Optical sensors such as GelSight~\cite{yuan2017gelsight} excel at capturing textures, contact shapes, and deformations~\cite{Akinola2025,ManiFeel}. Recent efforts to simulate tactile sensor deformation have focused on modeling the interaction between rigid objects and deformable elastomer surfaces. Traditional proxy-based approaches~\cite{Wedge,Donlon2018} often allow object–gel penetration as a surrogate for deformation. Although computationally efficient, these methods lack physical realism and fail to capture accurate stress–strain responses. Physics-based methods offer more fidelity but face practical challenges. Among them, finite element methods (FEM)~\cite{9561969} can deliver high accuracy in modeling large deformations; however, they require carefully constructed meshes and often demand adaptive remeshing strategies to avoid numerical instabilities at sharp contacts~\cite{TacIPC,8722757}. This mesh dependency hinders scalability and complicates simulation pipelines. On the other hand, the Material Point Method (MPM)~\cite{Tacchi} eliminates the need for explicit meshing and is robust to topological changes, but its accuracy depends heavily on particle density. High-resolution particle discretizations quickly lead to prohibitive memory-cost, making it difficult to capture complex surface wrinkles and contact details. Briefly, FEM achieves high fidelity but at the expense of meshing complexity, whereas MPM provides flexibility but is bottlenecked by memory consumption in high-resolution scenarios. 

To address this challenge, we propose a hybrid neural-physical simulation framework that enhances coarse MPM simulations with a learned, continuous representation of fine-scale deformation. Our key insight is that the high-dimensional state space of detailed tactile deformation can be effectively compressed into a low-dimensional latent manifold by principles of reduced-order modeling (ROM)~\cite{Lyu_2025,Reduced_ICRA2021}. We leverage this idea to train a neural field, an implicit neural representation that encodes the topology and deformation patterns of the elastomer from high-resolution MPM data. This neural field operates as a super-resolution decoder, mapping low-dimensional latent codes to high-fidelity deformation fields conditioned on the coarse simulation state.

\begin{figure}
    \centering
    \includegraphics[width=1.0\linewidth]{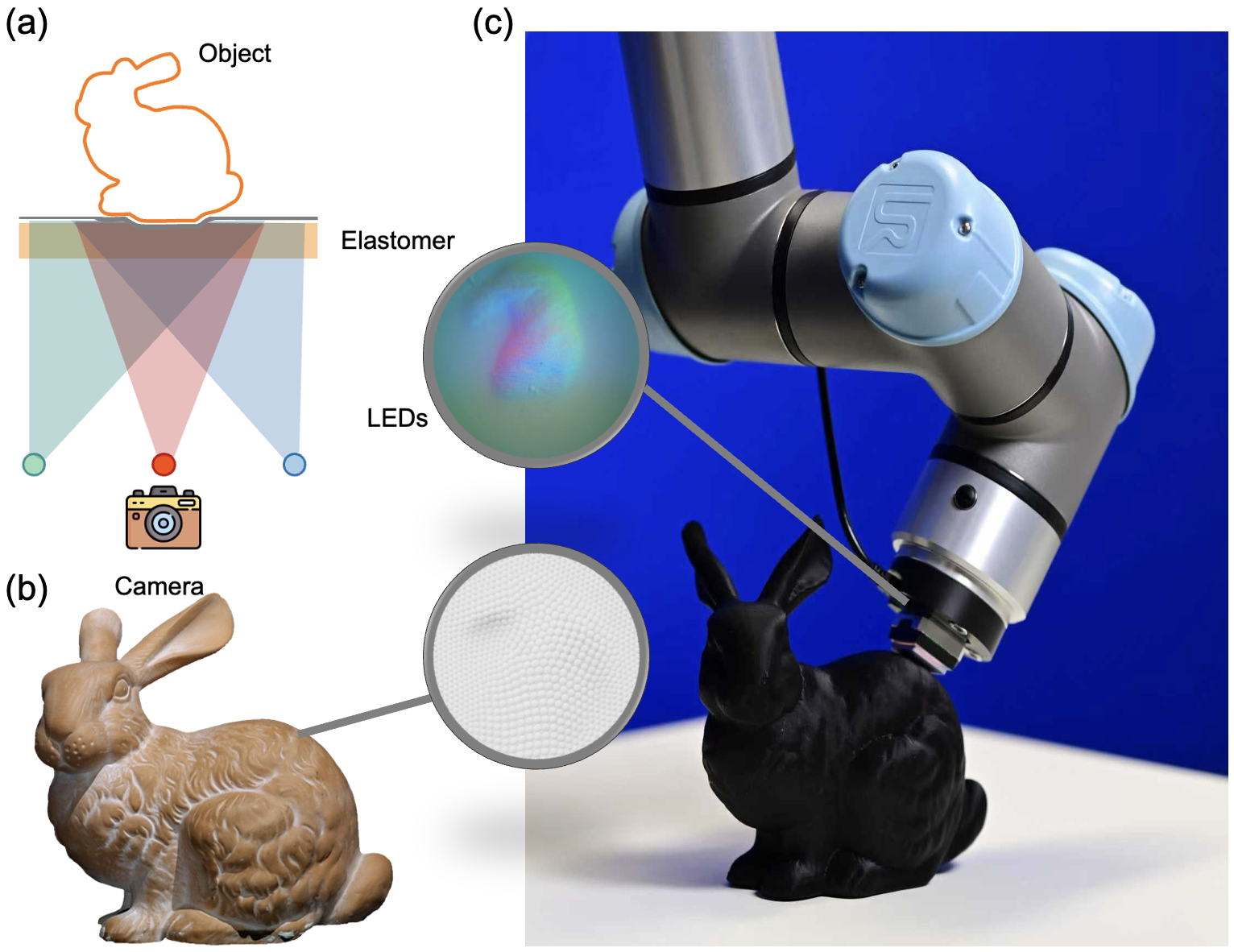}
    \caption{(a) The bunny presses against a soft elastomer, deforming its surface under illumination from multiple LEDs. The camera beneath captures modulated RGB reflections, encoding high-resolution 3D contact geometry.
(b) Our method reconstructs fine-scale surface details, such as hair-like textures on a 3D bunny model, directly from our reduced-order model, achieving high tactile geometric fidelity.
(c) Our physical experiment contains a UR5e robot with a GelSight Mini sensor that contacts the bunny, enabling direct validation of simulation accuracy against real-world tactile data.}
    \label{fig:teaser}
    \vspace{-10pt}
\end{figure}

During inference, our method decouples physical dynamics from detail reconstruction: the MPM solver handles global mechanics efficiently at low resolution, while the neural field refines the deformation field in efficient time, generating sub-particle-level tactile details. By querying the neural field directly on a dense grid, we effectively super-resolve the displacement and stress fields, recovering critical tactile features such as surface wrinkles, contact edges, typically smoothed out or missed entirely in low-resolution simulations. Importantly, our approach remains differentiable and preserves the underlying physics of MPM, ensuring stable and plausible interactions. The neural field is trained offline on paired high- and low-resolution MPM simulations, learning a mapping from coarse states to fine-grained deformation details. Once trained, it introduces minimal computational overhead during online execution, making it compatible with real-time robotic simulation and gradient-based optimization. This combination of physical grounding and expressive neural modeling enables both efficiency and high perceptual fidelity.

\begin{itemize}
\item We propose a {reduced-order neural simulation framework} that couples coarse MPM dynamics with an implicit neural decoder, enabling sub-particle tactile detail reconstruction from low-dimensional latent states while achieving significant computational efficiency.

\item We experimentally validate our framework against prior art through both simulation and real-robot experiments (Fig.\ref{fig:teaser}), demonstrating complex surface geometric fidelity in press simulation, tactile image rendering, and 3D surface reconstruction tasks with consistent sim-to-real transfer.
\end{itemize}

%% file: tex/Related_work.tex
\section{Related Work}
\begin{figure}[t]
    \centering
    \includegraphics[width=\linewidth]{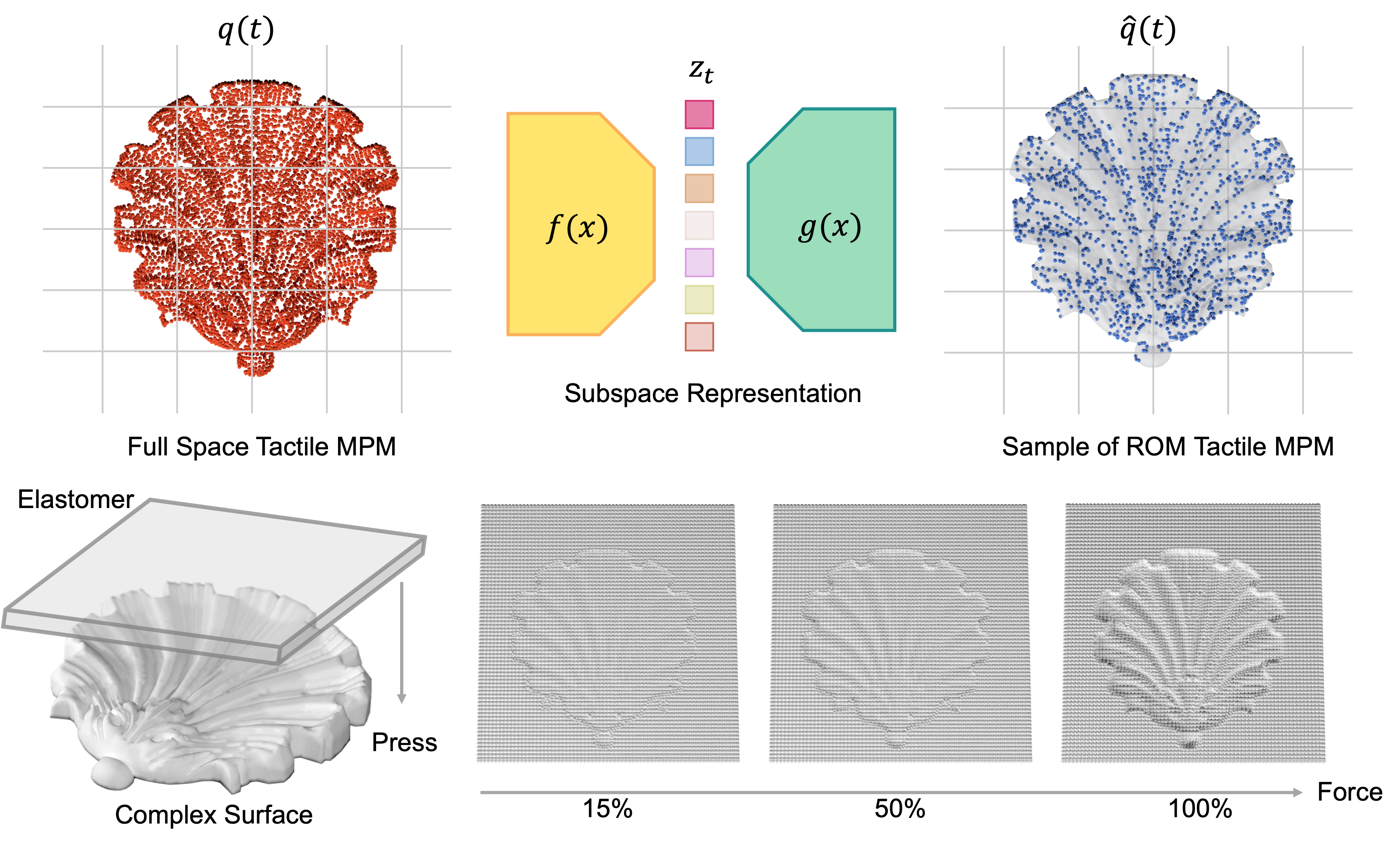}
    \caption{This figure illustrates our pipeline. The top row depicts the transformation from a high-dimensional full-space MPM simulation (left, red particles) to a low-dimensional subspace representation, and then back to a sample of the reduced-order model (ROM) MPM simulation. The bottom row demonstrates the tactile rendering process: when a complex surface is pressed against an elastomer, our method generates deformation maps at varying force levels.}
    \label{fig:pipeline}
\end{figure}
\subsection{Tactile Simulation}

The development of tactile simulation has followed a clear trajectory from rigid-body approximation to continuum and particle-based deformable modeling. \textit{Early attempts} relied on rigid-body simulators, where soft contact was approximated through geometric penetration or penalty-based models.  Prior art such as Tacto~\cite{TACTO} and Tactile-Gym~\cite{church2021optical} extended PyBullet~\cite{coumans2021} to render tactile images from contact geometry, while the efficient simulator of Xu~\emph{et al.}~\cite{xu2022efficient} incorporated a penalty-based formulation to estimate local force distribution.   Although computationally lightweight, these approaches approximate soft contact dynamics and thus sacrifice physical realism. \textit{Finite Element Method} (FEM)-based simulators \cite{Ma2019} are subsequently introduced to improve physical accuracy in soft material modeling. Representative works such as TacIPC~\cite{TacIPC} employ FEM and incremental potential contact formulations to capture detailed contact interactions and large deformation.  
Taxim~\cite{Taxim} further proposes a superposition strategy to approximate FEM responses while reducing computational cost.  
However, FEM-based approaches remain expensive and difficult to integrate into robot learning or real-time control pipelines due to remeshing and solver complexity. \textit{Material Point Method} (MPM)-based frameworks have recently emerged as a more efficient alternative for soft-body and tactile simulation.  MPM combines particle-based representation with a background grid, naturally supporting large deformation, topological change, and differentiable implementation. Tacchi~\cite{Tacchi} and Tacchi2.0~\cite{Tacchi2} exemplify this trend, providing lightweight, pluggable elastomer deformation simulators built on the Taichi~\cite{Taichi} programming framework. 

Our work builds upon this trajectory by coupling a differentiable MPM core with reduced-order neural decoding to achieve high-detail surface tactile perception without a large number of particles.





\subsection{Differentiable Simulation for Elastomer}

Differentiable simulation enables gradient-based optimization in soft elastomer by providing exact sensitivities of physical dynamics~\cite{Bacher2021}.  
Existing approaches can be broadly categorized into three types: (i) {Analytic methods} derive explicit gradients via sensitivity or adjoint analysis~\cite{bern2019}.  
They provide highly accurate derivatives for governing equations and material models, but require extensive manual derivation and symbolic differentiation for each new constitutive law or boundary condition.  
Such analytic formulations are often used in elastomer modeling where precise stress–strain gradients are needed for parameter calibration or inverse elasticity problem\cite{du2021}.  (ii) {Automatic differentiation solvers}~\cite{Taichi} leverage differentiable programming frameworks to propagate gradients through discretized physics pipelines.  
Systems such as PlasticineLab~\cite{huang2021plasticinelab}, employ moving least squares material point methods (MLS-MPM)~\cite{hu2018moving} to achieve differentiable soft-body dynamics, enabling joint optimization of control and morphology in complex deformable systems.  
These methods are general and flexible but often computationally heavy for dense elastomer simulations. (iii) {Neural surrogates}~\cite{sun2021amortized} learn an end-to-end mapping from system states to gradients or next-step dynamics, amortizing the cost of differentiation.  

In tactile sensing, differentiable simulators have proven particularly valuable for solving the elastomer deformation problem~\cite{si2024difftactile,9829271}.  
By directly comparing simulated and observed deformations, these models enable gradient-based recovery of contact geometry, force distribution, or material parameters that are ill-posed for traditional numerical solvers.  However, high-fidelity differentiable elastomer simulations remain computationally expensive due to the need for dense particle or mesh resolutions.  

Our framework addresses this limitation by combining differentiable analytic simulation with reduced-order modeling, thereby achieving physical consistency at a minimal cost.

\subsection{Reduced-order Neural Fields Modeling}

Reduced-order modeling (ROM) accelerates simulation by projecting dynamics onto a low-dimensional manifold. Early methods used linear subspaces from modal analysis\cite{1308087} or piecewise control~\cite{Reduced_ICRA2021,10341432}, but struggle with nonlinear deformations. Recent approaches employ autoencoder-based neural networks ~\cite{Fulton2019, shen2021} and implicit neural representations~\cite{zong2023, chen2023} to learn richer, nonlinear reduced spaces. However, excessive nonlinearity can lead to overfitting and slow convergence~\cite{sharp2023} issues that hybrid linear-neural architectures can help mitigate~\cite{shen2021}.

In this work, we combine autoencoder-based dimensionality reduction with differentiable latent-space dynamics for high-detail tactile perception. To improve efficiency, we use approximate forces via sparse spatial sampling during training. This efficiently enforces smoothness and enables the stable learning of deformation models, allowing for fine-grained contact interaction.

%% file: tex/Pipeline_Overview.tex
\section{Methods}

\subsection*{Preliminaries: MPM in Tactile Simulation}

The Material Point Method (MPM) discretizes elastomer materials as a set of particles that carry mass, velocity, and deformation gradients, which interact through a background Eulerian grid~\cite{Tacchi}. 
At each time step, particle mass and momentum are transferred to grid nodes (P2G), grid velocities are updated under internal and external forces, and results are interpolated back to particles (G2P). 
The deformation gradient of each particle evolves as
\begin{equation}
\begin{split}
    \mathbf{F}_i^{t+1} 
    &= (\mathbf{I} + \Delta t\, \nabla \mathbf{v}_i)\, \mathbf{F}_i^t, \\
    \nabla \mathbf{v}_i 
    &= \sum_{\mathbf{g} \in \mathcal{G}} 
       \mathbf{v}_{\mathbf{g}} \otimes \nabla w(\mathbf{x}_i - \mathbf{x}_{\mathbf{g}}),
\end{split}
\label{eq:mpm_update}
\end{equation}
where $\mathbf{F}_i$ is the deformation gradient of particle $i$, 
$\mathbf{v}_{\mathbf{g}}$ denotes the grid node velocity, 
$w(\cdot)$ is the quadratic B-spline weighting kernel, 
and $\nabla w(\cdot)$ is its spatial gradient that determines particle–grid coupling.
We adopt the affine particle-in-cell (APIC) variant~\cite{affine} to improve stability and angular momentum conservation. The details refer to Tacchi\cite{Tacchi}.

While MPM robustly captures elastomer dynamics and contact interactions, its computational cost scales linearly with the number of particles $N$. 
Accurately modeling complex surface deformations (e.g., wrinkles and micro-indentations) typically requires millions of particles, rendering high-fidelity tactile simulation computationally prohibitive.

\begin{figure}
    \centering
    \includegraphics[width=\linewidth]{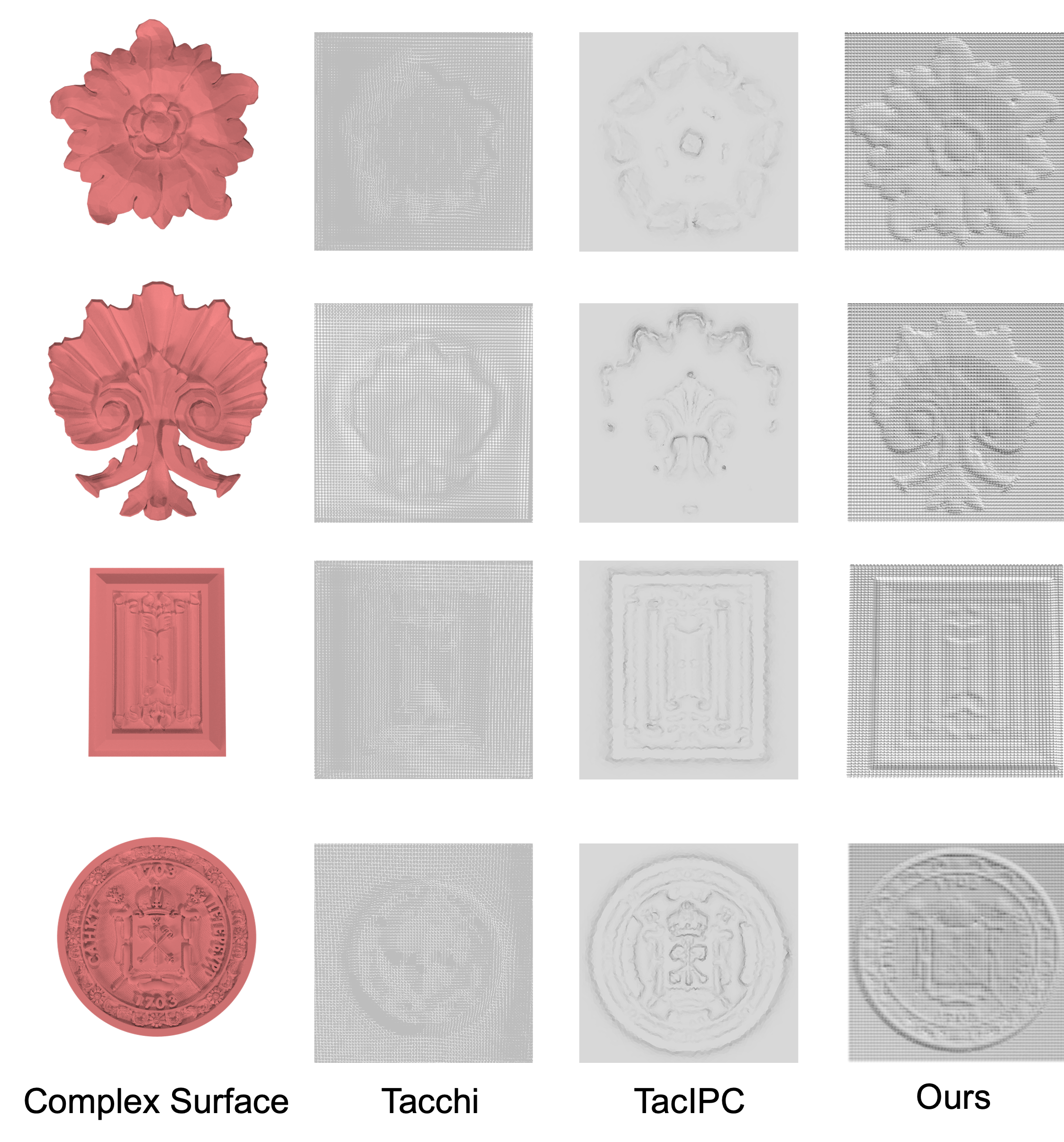}
    \caption{Comparison of Gelsight press simulation results. We present an evaluation of Tacchi\cite{Tacchi}, TacIPC\cite{TacIPC}, and our proposed method under identical pressure conditions and equivalent element counts, while the red model is the input complex surface. Our simulation results demonstrate superior fidelity in capturing the complex surface morphology, outperforming previous work. }
    \label{fig:compare_ICRA}
\end{figure}

\subsection{Reduced-order Neural Modeling}
High-resolution tactile simulation incurs prohibitive computational cost due to the large number of particles; to this end, we reformulate the problem in a reduced-order setting.

Let the full particle-based configuration at time step $ t $ be
\begin{equation}
    \mathbf{q}_t = \left\{ \mathbf{x}_i^t, \mathbf{F}_i^t \right\}_{i=1}^{N} \in \mathbb{R}^n,
\end{equation}
where $ \mathbf{x}_i^t \in \mathbb{R}^3 $ and $ \mathbf{F}_i^t \in \mathbb{R}^{3 \times 3} $ denote the position and deformation gradient of particle $ i $, respectively. Standard MPM evolves $ \mathbf{q}_t $ through particle-to-grid transfers, grid updates, and grid-to-particle advections, as described in the preliminaries. However, the computational complexity scales with $ n $, which becomes intractable for $ n \sim 10^6 $.

We therefore introduce a nonlinear subspace parameterization:
\begin{equation}
    \mathbf{q}_t \approx f_\theta(\mathbf{z}_t), \quad \mathbf{z}_t \in \mathbb{R}^r, \quad r \ll n,
\end{equation}
where $ f_\theta: \mathbb{R}^r \to \mathbb{R}^n $ is a neural decoder that reconstructs the full particle states from a compact latent coordinate $ \mathbf{z}_t $. Instead of evolving $ \mathbf{q}_t $ directly, we solve for $ \mathbf{z}_t $ such that the governing MPM dynamics are satisfied in the reduced subspace. The latent update is obtained by minimizing the energy functional projected onto the subspace manifold:
\begin{equation}
    \mathbf{z}_{t+1} = \arg\min_{\mathbf{z}} \left[ 
        \frac{1}{2 \Delta t^2} \| f_\theta(\mathbf{z}) - \mathbf{q}_{t+1}^{\text{inertial}} \|_M^2 + P(f_\theta(\mathbf{z}))
    \right],
    \label{eq:latent_update}
\end{equation}
In the reduced-order formulation, the implicit inertial penalty term is projected onto a latent subspace, resulting in \(\frac{1}{2\Delta t^2} \big\| f_\theta(z) - q_{t+1}^{\text{inertial}} \big\|_{\mathbf{M}}^2\), where \(\Delta t^2\) denotes the square of the time step, consistent with the second-order temporal discretization used in the physical integration scheme.
While $ \mathbf{q}_{t+1}^{\text{inertial}} $ is the inertial guess propagated from the previous state, $ \|\cdot\|_M $ denotes the mass-weighted norm induced by the particle mass matrix $ \mathbf{M} $, and $ P(\cdot) $ denotes the total potential energy, including both internal elastic and external loading.

This formulation enables latent dynamics to faithfully approximate the high-dimensional evolution while operating in a much smaller dimensional space.

\subsection{Subspace Representation for High-Detail Tactile}

The effectiveness of reduced-order simulation hinges on the expressiveness of the subspace manifold. In tactile elastomer modeling, fine-grained contact features, such as ridges, wrinkles, and localized stress fields are critical for perceptual fidelity but are easily lost under naive dimensionality reduction. To preserve these details, we construct the subspace using a neural autoencoder trained to capture high-resolution deformation modes (see Fig.\ref{fig:pipeline}). We observe that our method effectively reduces the particle dimensionality while preserving sharp geometric features, as highlighted by the blue points (e.g., edges and protrusions).

Specifically, we employ an autoencoder structure $ (f_\theta, g_\theta) $, where the encoder $ f_\theta: \mathbb{R}^n \to \mathbb{R}^r $ maps full particle configurations to latent coordinates, and the decoder $ g_\theta: \mathbb{R}^r \to \mathbb{R}^n $ reconstructs them:
\begin{align}
    \mathbf{z}_t &= f_\theta(\mathbf{q}_t), \\
    \hat{\mathbf{q}}_t &= g_\theta(\mathbf{z}_t).
\end{align}

The decoder $ g_\theta $ serves as the reduced-order mapping used during simulation. Training is conducted on paired low- and high-resolution MPM trajectories: the reconstruction loss enforces that high-resolution particle states are recoverable from their latent codes.

To ensure physical consistency between the reduced latent dynamics and the coarse-grained MPM updates, we introduce a consistency loss that penalizes deviations in momentum and deformation evolution. 
Specifically, given the coarse MPM prediction $\tilde{\mathbf{q}}_{t}$ obtained from the low-resolution simulation and the reconstructed high-resolution state $\hat{\mathbf{q}}_{t}=g_\theta(f_\theta({\mathbf{q}}_{t}))$, the loss is defined as
\begin{equation}
\mathcal{L}_{\text{cons}} =
\lambda_v \sum_{i\in\mathcal{P}} 
    \|\,\hat{\mathbf{v}}_i^{\,t} - {\mathbf{v}}_i^{\,t}\|^2
+ \lambda_F \sum_{i\in\mathcal{P}}
    \|\hat{\mathbf{F}}_i^{\,t} - {\mathbf{F}}_i^{\,t}\|^2 ,
\label{eq:consistency_loss}
\end{equation}
where $\hat{\mathbf{v}}_i^{\,t}$ and $\hat{\mathbf{F}}_i^{\,t}$ are the velocity and deformation gradient of particle $i$ reconstructed from the latent code, while ${\mathbf{v}}_i^{\,t}$ and ${\mathbf{F}}_i^{\,t}$ are their counterparts computed by the coarse MPM solver. The coefficients $\lambda_v$ and $\lambda_F$ balance momentum and strain consistency, with $\lambda_v = 0.4$ and $\lambda_F = 0.6$ in our setting. This loss term enforces that the reduced representation evolves coherently with the underlying physics, preserving local stress evolution and global momentum conservation during training.

Once trained, the decoder implicitly encodes a nonlinear subspace that spans both global deformation patterns and localized high-frequency details. During simulation, solving for $ \mathbf{z}_t $ via the reduced optimization in Eq.~\eqref{eq:latent_update} yields high-detail tactile surface reconstructions at a fraction of the original computational cost. The recovered high-resolution configuration $\hat{\mathbf{q}}_t = g_\theta(\mathbf{z}_t)$
is then used for tactile rendering, enabling detailed depth map generation without the overhead of full-resolution MPM simulations.

\begin{figure}
    \centering
    \includegraphics[width=\linewidth]{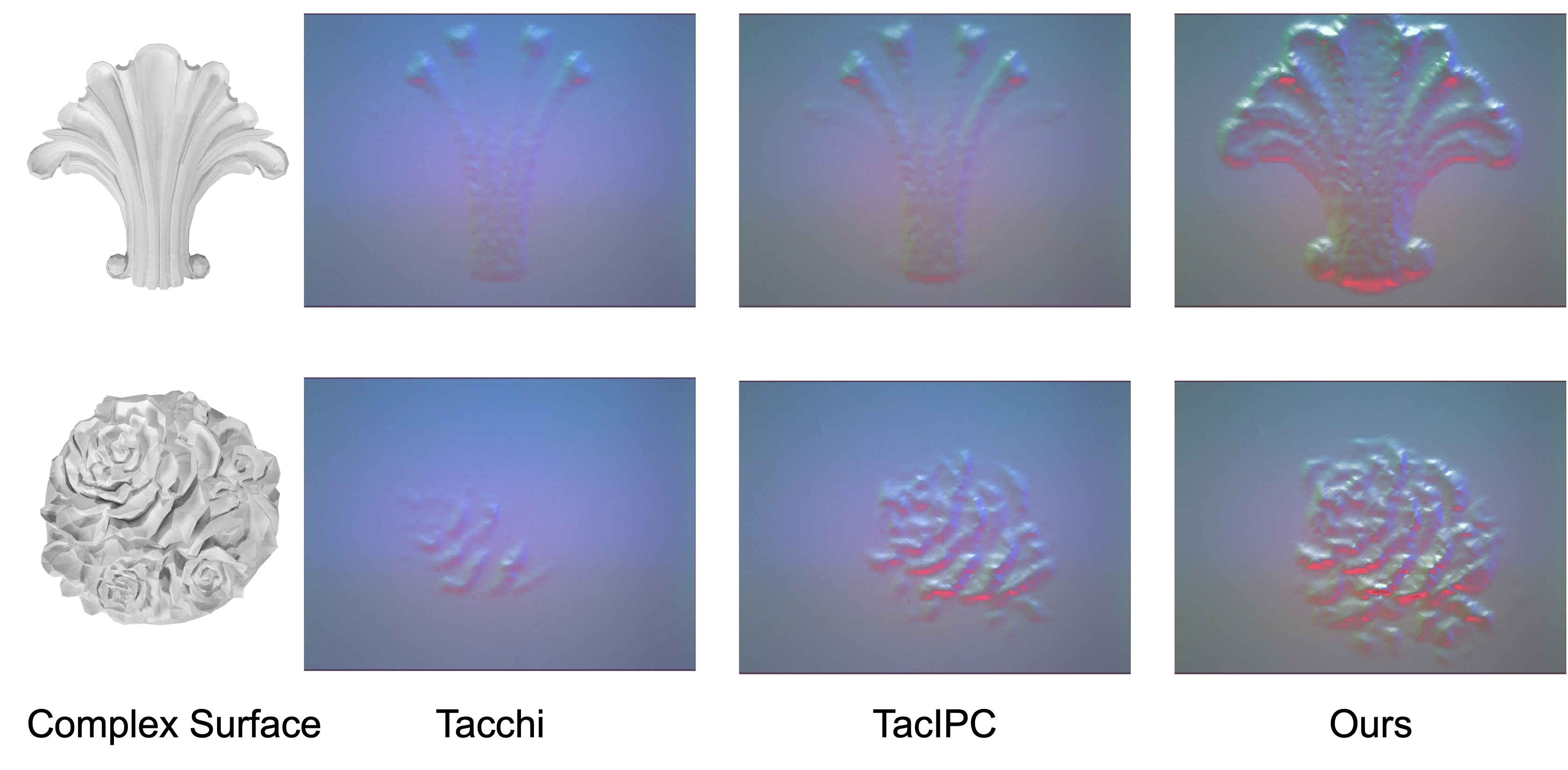}
    \caption{We denote the input complex surface geometry in white and render tactile depth maps using Tacchi, TacIPC, and our reduced-order neural method. While Tacchi\cite{Tacchi} and TacIPC\cite{TacIPC} suffer from smoothing artifacts and loss of fine-scale features (e.g., wrinkles, ridges), our approach preserves high fidelity details, even from a low-dimensional latent state. Crucially, it generates realistic tactile images directly via learned geometric reconstruction, eliminating the need for full-resolution MPM simulation.
}
    \label{fig:rendering}
\end{figure}

\subsection{Tactile Image Rendering from Reduced States}
\label{Rendering}

In MPM-based tactile simulation, the tactile signal is captured through a camera observing the elastomer surface through a transparent gel layer. 
Instead of a direct orthographic projection, the apparent depth map $D_t(u)$ is computed by tracing view rays refracted through the gel and intersecting them with the deformed elastomer surface. 
We first define the reconstructed surface particle set as
\begin{equation}
    \hat{\mathcal{S}}_t = 
    \left\{ \hat{\mathbf{x}}_i^t \,\middle|\, i \in \mathcal{P}_{\text{surface}} \right\},
    \label{eq:surface_set}
\end{equation}
where $\hat{\mathcal{S}}_t$ denotes the reconstructed surface configuration obtained from the decoder $g_\theta(\mathbf{z}_t)$, and $\mathcal{P}_{\text{surface}}$ indexes particles on the elastomer–object contact boundary. 

Each surface point $\hat{\mathbf{x}}_i^t$ is projected onto the sensor plane along a refracted viewing direction $\mathbf{d}_i$, which accounts for the bending of light at the gel–air interface with normal $\mathbf{n}_0$ and refractive index ratio. 
The resulting depth value is given by
\begin{equation}
    D_t(u) = \langle \mathbf{n}_s, \hat{\mathbf{x}}_i^t \rangle 
    + \frac{z_s - \langle \mathbf{n}_s, \hat{\mathbf{x}}_i^t \rangle}
           {\langle \mathbf{n}_s, \mathbf{d}_i \rangle},
    \label{eq:depth_refr}
\end{equation}
where the denominator $\langle \mathbf{n}_s, \mathbf{d}_i \rangle$ accounts for the angular deviation introduced by refraction. Here, $z_s$ denotes the signed distance of the sensor plane along its normal direction $\mathbf{n}_s$, serving as the reference depth at which the tactile image is formed.  When $\mathbf{d}_i$ aligns with the sensor normal $\mathbf{n}_s$ (i.e., no refraction), naturally degenerates into the orthographic projection $D_t(u) = \langle \mathbf{n}_s, \hat{\mathbf{x}}_i^t \rangle$.

This rendering formulation ensures that geometric deformation and optical projection are consistently modeled in the same physical framework (as shown in Fig.\ref{fig:rendering}). 
Given the latent state $\mathbf{z}_t$ from the reduced-order dynamics, the decoder reconstructs the high-resolution surface $\hat{\mathcal{S}}_t$, and substituting $\hat{\mathbf{x}}_i^t \in \hat{\mathcal{S}}_t$ into Eq.~\eqref{eq:depth_refr} yields the tactile depth map value directly from the reduced dynamics. 
As a result, tactile image formation becomes a physically grounded outcome of the reduced-order simulation, capturing fine-scale geometric and photometric variations without full-resolution MPM computation.

%% file: tex/Experiments.tex
\section{Experiments}
\subsection{Implementation Details}
\label{sec:details}

Our simulations and training are implemented in Taichi~\cite{Taichi} with PyTorch for neural modules. The reduced-order autoencoder consists of an MLP encoder-decoder pair with three hidden layers of sizes $[512, 256, 128]$ and latent dimension $r = 64$ (which work well in initial trials), each followed by ReLU activations and layer normalization. The decoder output reconstructs per-particle displacement and deformation gradient. Training data are generated from paired high- and low-resolution MPM simulations of $10^6$ and $10^4$ particles, respectively, using identical boundary conditions and materials. Each sequence contains quasi-static indentation with varying force levels, yielding approximately 6k training pairs. Networks are optimized with Adam ($\text{lr} = 1 \times 10^{-4}$, $\beta_1 = 0.9$, $\beta_2 = 0.999$) for 3k epochs on RTX 3090 GPU. Each training epoch takes $\approx 7$~s with a mini-batch size of 32, and the model converges in $\sim 12$~h wall-clock time. At inference, coarse MPM dynamics run with $N_\text{coarse} = 10^4$ particles on a $100 \times 100 \times 21$ grid. Latent optimization per frame (Eq.~4) is solved using L-BFGS for at most 10 iterations (tolerance $10^{-5}$).

\vspace{-5mm}
\subsection{Experimental Setup}
\label{sec:Setup}
To validate our method under realistic robotic setup, we use a UR5e robotic arm equipped with a GelSight Mini optical tactile sensor. Following the capture policy of prior work~\cite{Tacchi}, we fabricate eight custom FDM-printed objects with complex surface geometries. Each object is pressed at nine locations arranged in a $3\times3$ grid (8\,mm spacing), with indentation depths ranging from 0 to 0.5\,mm in 0.1\,mm steps, yielding high-resolution tactile image sequences under controlled lighting and motion. This robotically collected dataset serves as a real-world benchmark for evaluating physical fidelity and generalization. 

For simulation evaluation, we compare against Tacchi~\cite{Tacchi} and TacIPC~\cite{TacIPC}. By decoupling coarse MPM dynamics from neural super-resolution, our method recovers sub-particle tactile details without mesh constraints, achieving TacIPC-level geometric fidelity at significantly faster inference speed.

\begin{figure}
    \centering
    \includegraphics[width=\linewidth]{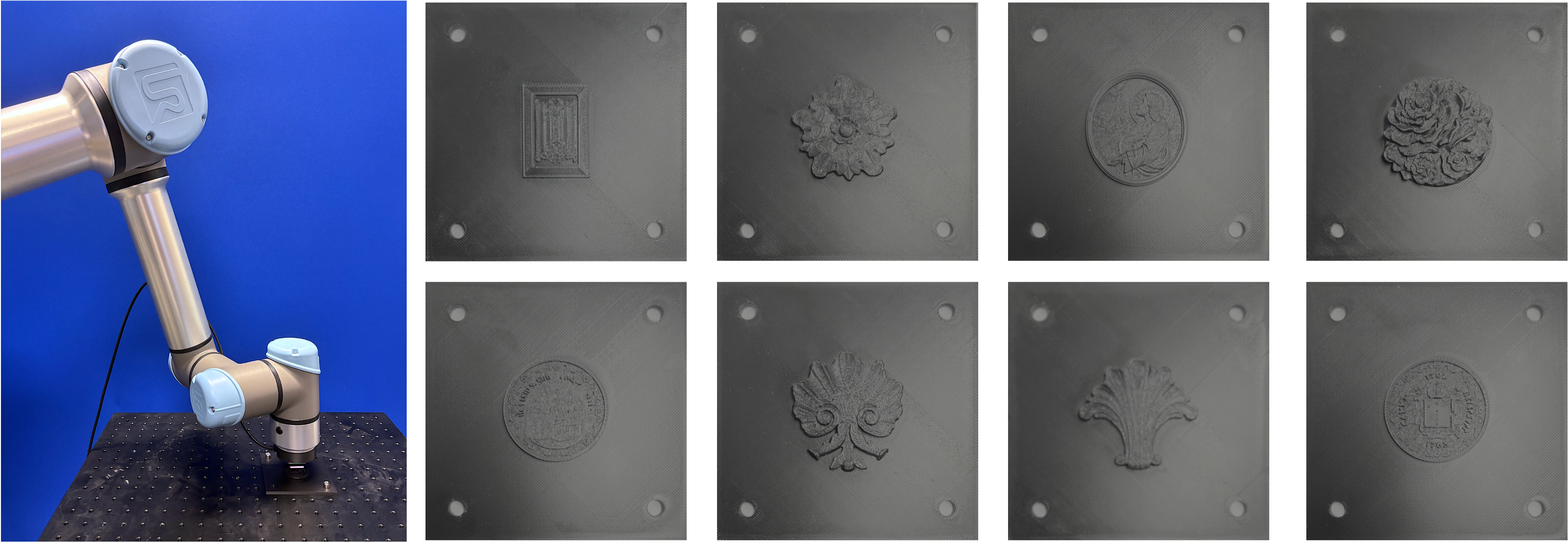}
        \caption{\textit{Left:} UR5e robotic arm equipped with a GelSight Mini sensor for planar tactile perception. The sensor is mounted on the end-effector, ready to interact with objects fixed on the workbench. \textit{Right:} Our custom-designed dataset of eight 3D-printed models featuring complex, non-trivial surface geometries (including intricate curvatures, localized indentations, and asymmetric topographies).}
    \label{fig:dataset_gallery}
\end{figure}

\begin{table}[t]
\centering
\caption{Pseudo-image quality evaluation.}
\label{tab:quality_comparison}
\begin{tabularx}{\columnwidth}{lXXXX}
\hline
Method & SSIM$\uparrow$ & MAE$\downarrow$ & PSNR$\uparrow$ & Rendering Time(s) $\downarrow$ \\
\hline
Tacchi\cite{Tacchi}       & 0.880 & 0.034 & 55.4 & 0.38 \\
TacIPC\cite{TacIPC}        & 0.902 & 0.037 & 57.7 & 0.25 \\
Ours          & \textbf{0.923} & \textbf{0.028} & \textbf{61.5} & \textbf{0.17} \\
\hline
\end{tabularx}
\end{table}

\subsection{Press Simulation}

To enable direct comparison with Tacchi~\cite{Tacchi} and TacIPC~\cite{TacIPC}, we simulate a $30 \,\text{mm} \times 30 \,\text{mm} \times 4 \,\text{mm}$ silicone elastomer with Young's modulus $E = 1.19 \times 10^5 \,\text{Pa}$ and Poisson's ratio $\nu = 0.43$ following empirical calibration~\cite{TacIPC}. We employ a coarse MPM solver with $N_{\text{coarse}} = 10^{4}$ particles on a $100 \times 100 \times 21$ grid, which captures global deformation patterns efficiently but fails to preserve microscale features such as wrinkles, asymmetric local textures, and small contact discontinuities.  Our reduced-order neural framework addresses this limitation by super-resolving the latent state $\mathbf{z}_t$ into a high-fidelity displacement field equivalent to over $10^6$ particles. This reconstruction not only recovers the lost geometric fidelity but also provides stable gradients for downstream optimization tasks. In contrast, Tacchi’s memory-cost displacement maps suffer from partial surface collapse in regions of high curvature, while TacIPC’s FEM-based outputs remain bounded by mesh resolution, producing visible interpolation artifacts in contact zones. As shown in Fig.~\ref{fig:compare_ICRA}, our reconstructed tactile geometry closely matches the ground-truth reference (red surface), successfully reproducing microscopic ridges, localized indentations, and nonlinear surface undulations that are absent in both baseline methods. 

\subsection{Rendering Evaluation}
Our formulation preserves fine-scale surface details without full-resolution MPM simulation, showing that tactile rendering emerges directly from the reduced-order latent dynamics rather than post-processing. By coupling latent evolution with surface reconstruction, our framework synthesizes high-fidelity tactile images efficiently.

We evaluate rendering fidelity against real GelSight data and prior simulators~\cite{TacIPC,Tacchi} using SSIM, MAE, and PSNR. As shown in Tab.~\ref{tab:quality_comparison}, our method consistently outperforms baselines, improving SSIM and PSNR while reducing MAE by over 20\%, with approximately 50\% lower inference time per frame. These results demonstrate superior geometric and photometric fidelity at higher efficiency.

Qualitative comparisons (Fig.~\ref{fig:rendering}) further show that, unlike Tacchi and TacIPC which oversmooth high-frequency features, our method preserves sharp local details and complex contact geometry, enabled by rendering directly from reduced latent states (Sec.~\ref{Rendering}).

\vspace{-4pt}
\begin{figure}[t]
    \centering
    \includegraphics[width=\linewidth]{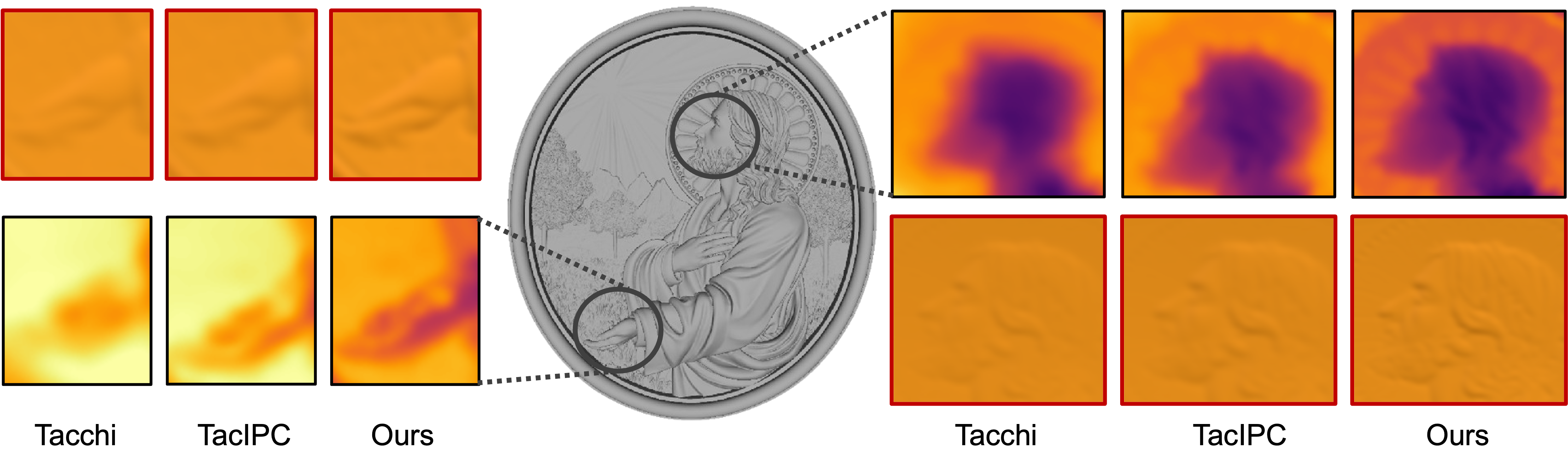}
    \vspace{-6pt}
    \caption{Comparison of depth prediction quality across simulators. Each tactile simulator generates tactile images under identical contact poses.}
    \vspace{-4pt}
    \label{fig:recon}
\end{figure}
\vspace{-8pt}

\subsection{Tactile  Reconstruction}
\label{sec:tactile-mesh-recon}

While previous tactile simulators mainly generate 2D pseudo-images, physically accurate 3D surface reconstruction is essential for tasks that require geometric or force reasoning, such as grasp stability analysis or system identification. We therefore conduct a fully simulated evaluation that measures how well the deformed elastomer surface can be reconstructed from rendered tactile images, following the protocol in~\cite{TacIPC}. A U-Net is trained to predict depth maps from   simulated tactile images produced by corresponding simulators. The predicted depth maps are back-projected to 3D and converted into meshes for comparison with the original surfaces from simulation.

We evaluate reconstruction accuracy using Chamfer distance. Our method achieves the lowest error, outperforming Tacchi’s over-smoothed results and TacIPC’s loss of fine contact details. The implicit neural decoder preserves sharp edges and micro-wrinkles while maintaining physical plausibility. As shown in Fig.~\ref{fig:recon}, both predicted depth and reconstructed meshes remain structurally coherent and free of depth-induced artifacts, indicating physically consistent and recoverable 3D geometry.

\subsection{Computational Cost}
\label{sec:cost}

Our framework achieves high computational efficiency by decoupling coarse physical simulation from fine geometric reconstruction. Global dynamics are solved with a coarse MPM, while fine surface details are recovered through lightweight neural inference, avoiding the cost of simulating millions of particles or dense FEM meshes.

In practice, full-resolution MPM with $>10^6$ particles requires large GPU memory, while FEM+IPC~\cite{TacIPC} offers comparable fidelity but remains computationally expensive (about 3.6 min on average) due to implicit solving and remeshing overhead.
Tacchi~\cite{Tacchi}, though lighter, relies on memory-intensive precomputed maps $(\sim 5~\text{GB})$ that scale poorly with resolution.
In contrast, our reduced-order latent model reconstructs high-fidelity surfaces at only 1.20 min  on a single RTX 3090 GPU, about 3× faster than TacIPC and over 4.3× faster than Tacchi, while maintaining nearly the same Chamfer distance as full MPM.
\begin{table}[t]
\centering
\caption{\textbf{Runtime, memory, and accuracy comparison.}
Averages of the same tactile interaction.}
\label{tab:runtime_ablation}
\vspace{4pt}
\small
\setlength{\tabcolsep}{4pt}
\begin{tabular}{@{}lccc@{}}
\toprule
Method & Time [min] $\downarrow$ & Mem [GB] $\downarrow$ & Chamfer-$L_2$ $\downarrow$ \\
\midrule
Full MPM ($10^6$ p) & 6.83 & 10.4 & 0.34 \\
FEM+IPC~\cite{TacIPC} & 3.57 & 3.9 & 0.39 \\
Tacchi~\cite{Tacchi} & 5.21 & 5.1 & 0.43 \\
\midrule
\textbf{Ours (full)} & \textbf{1.20} & \textbf{3.5} & \textbf{0.35} \\
\quad w/o physics & 0.76 & 1.3 & 0.45 \\
\quad w/o multi-scale & 0.98 & 1.4 & 0.40 \\
\quad latent dim $r{=}16$ & 1.05 & 1.7 & 0.43 \\
\bottomrule
\end{tabular}
\vspace{-4pt}
\end{table}

As summarized in Tab.~\ref{tab:runtime_ablation}, our method attains the best trade-off between runtime, memory footprint, and geometric accuracy. 
Removing physics-consistency constraints accelerates inference slightly but causes noticeable geometric degradation and unstable contact behavior. 
Similarly, removing multi-scale supervision leads to oversmoothed ridges and a $\sim$25\% increase in Chamfer error.
Reducing the latent dimension from $r\!=\!64$ to $r\!=\!16$ lowers runtime to $1.05$\,min average simulation, but at the cost of high-frequency detail loss, confirming that our chosen latent width provides a balanced compromise between efficiency and fidelity. This efficiency is achieved without sacrificing accuracy: as shown in Tab.~\ref{tab:runtime_ablation}, our method matches or exceeds the geometric fidelity of TacIPC while remaining significantly faster.

%% file: tex/secConclusion.tex
\section{Discussion and Conclusion }\label{sec:conclusion}
%
In this work, we presented a reduced-order neural simulation framework that couples coarse-grained MPM dynamics with an implicit neural decoder for high-detail tactile perception. By learning a compact latent manifold of elastomer deformation, our method reconstructs sub-particle tactile geometry with high fidelity while achieving substantial computational and memory savings. The framework remains physically grounded and fully differentiable, enabling efficient tactile rendering and geometry reconstruction that approach the accuracy of full-resolution MPM or FEM solvers at a fraction of their cost. Our current formulation targets quasi-static or low-speed indentation-dominated contact regimes. Highly dynamic interactions such as fast sliding, rolling, or impact introduce inertial effects and contact discontinuities that are not explicitly modeled in the present latent dynamics. In addition, generalization to unseen materials is limited, as elastomer deformation is highly sensitive to material parameters and our decoder is trained under fixed constitutive settings. We also focus on geometric fidelity of tactile deformation rather than absolute force or stress estimation, which typically requires precise material calibration and friction modeling. Finally, the tactile rendering assumes fixed illumination, which may limit photometric realism. Future work will explore incorporating isosurface-based contact representations~\cite{GaussianSlicer} and dynamic sliding interactions~\cite{Proprioceptive,10285446} to support richer contact behaviors and internal deformation inference~\cite{Design}, as well as extending the differentiable pipeline toward real-time tactile display~\cite{display}.

\section*{Acknowledgment}
This work was supported in part by the Hong Kong SAR Research Grants Council Early Career Scheme (RGC-ECS) under Grant CUHK/24204924, and University of Melbourne Early Career Researcher Grant PRJ\_026163.